\documentclass{article}

\usepackage[nonatbib, preprint]{neurips_2023}

\usepackage{hyperref}
\usepackage{url}
\usepackage{amsthm}
\usepackage{amsmath}
\usepackage[pdftex]{graphicx}
\usepackage{wrapfig}
\usepackage{algorithm}
\usepackage{algpseudocode} 

\theoremstyle{definition}

\usepackage[utf8]{inputenc} % allow utf-8 input
\usepackage[T1]{fontenc}    % use 8-bit T1 fonts
\usepackage{hyperref}       % hyperlinks
\usepackage{url}            % simple URL typesetting
\usepackage{booktabs}       % professional-quality tables
\usepackage{amsfonts}       % blackboard math symbols
\usepackage{nicefrac}       % compact symbols for 1/2, etc.
\usepackage{microtype}      % microtypography
\usepackage{xcolor}         % colors

\title{Multitask Extension of \\ Geometrically Aligned Transfer Encoder}

\author{%
  Sung Moon Ko$^*$ \\
  LG AI Research\\
  \texttt{sungmoon.ko@lgresearch.ai}\\
  \And
  Sumin Lee$^*$\\
  LG AI Research\\
  \texttt{sumin.lee@lgresearch.ai}\\
  \And
  Dae-Woong Jeong$^*$\\
  LG AI Research\\
  \texttt{dw.jeong@lgresearch.ai}\\
  \And
  Hyunseung Kim\\
  LG AI Research\\
  \texttt{hyunseung.kim@lgresearch.ai}\\
  \And
  Chanhui Lee\\
  LG AI Research\\
  \texttt{chanhui-lee@lgresearch.ai}\\
  \And
  Soorin Yim\\
  LG AI Research\\
  \texttt{soorin.yim@lgresearch.ai}\\
  \And
  Sehui Han\\
  LG AI Research\\
  \texttt{hansse.han@lgresearch.ai}\\
}

\begin{document}

\maketitle
\def\thefootnote{*}\footnotetext{These authors contributed equally to this work}\def\thefootnote{\arabic{footnote}}

\begin{abstract}
Molecular datasets often suffer from a lack of data. It is well-known that gathering data is difficult due to the complexity of experimentation or simulation involved. Here, we leverage mutual information across different tasks in molecular data to address this issue. We extend an algorithm that utilizes the geometric characteristics of the encoding space, known as the Geometrically Aligned Transfer Encoder (GATE), to a multi-task setup. Thus, we connect multiple molecular tasks by aligning the curved coordinates onto locally flat coordinates, ensuring the flow of information from source tasks to support performance on target data.
\end{abstract}

%%%%%%%%%%%%%%%%%%%%%%%%%%%%%%%%%%%%%%%%%%%%%%%%%%%%%%%%%%%%%%%%%%%%%%%%%%%%
%%%%%%%%%%%%%%%%%%%%%%%%%%%%%%%%%%%%%%%%%%%%%%%%%%%%%%%%%%%%%%%%%%%%%%%%%%%%
%%%%%%%%%%%%%%%%%%%%%%%%%%%%%%%%%%%%%%%%%%%%%%%%%%%%%%%%%%%%%%%%%%%%%%%%%%%%

\section{Introduction}
The quantity of data is a crucial factor in machine learning. However, it is not always feasible to acquire the necessary amount of data in practice. Many efforts have been made to address the data issue. One direct approach is data generation, which aims to generate plausible data (such as through reference augmentations or generation). Another approach is transfer learning, which is more indirect as it leverages mutual information from different source tasks \cite{https://doi.org/10.1002/sam.10099, doi:10.1137/1.9781611972825.47, 10.1145/2433396.2433449, 6606822, 9051683, Quattoni, Kulis2011WhatYS, DBLP:journals/corr/abs-1902-07208, YU2022230, wang2019, doi:10.1073/pnas.2024383118}. Lastly, there is multi-task learning, which shares a latent space across given tasks (see references on MTL).

Despite these achievements, the data issue remains particularly pronounced in scientific endeavors. Scientific experiments or simulations often require significant amounts of time and effort, making it challenging to amass abundant data in the field. However, our main focus is on molecular property prediction tasks \cite{scarselli2009, bruna2013, duvenaud2015, defferrard2016, jin2018, C8SC04228D, ko2023grouping}. We aim to address this issue by utilizing various molecular property datasets.

Our starting point is a transfer algorithm, namely the Geometrically Aligned Transfer Encoder (GATE), which is based on differential geometry \cite{ko2023geometrically}. This algorithm utilizes the concept of curved geometry in a Riemannian scheme. The key idea of this algorithm is to align the geometrical shapes of the underlying latent spaces of source and target tasks. In general, it is extremely complicated to compute their geometrical characteristics analytically. However, the algorithm bypasses this issue by introducing one crucial mathematical characteristic of Riemannian geometry: diffeomorphism invariance, which guarantees the freedom of coordinate choices at any point on the manifold. Additionally, it ensures that one can always find a locally flat frame under any circumstances. If one can find a locally flat frame over any tasks, then it is possible to require a constraint that restricts the geometric shape of coordinates over source and target tasks. If the underlying geometry can be matched, the mutual information across tasks will flow to one another and support model performance on the target task side. However, GATE is proven to work in a two-task setting, with one target and one source task. Yet, theoretically, it is not restricted to two tasks. Therefore, we extend the concept of GATE to multiple sources.

\begin{figure}[t!]
\begin{center}
\includegraphics[width=1\linewidth]{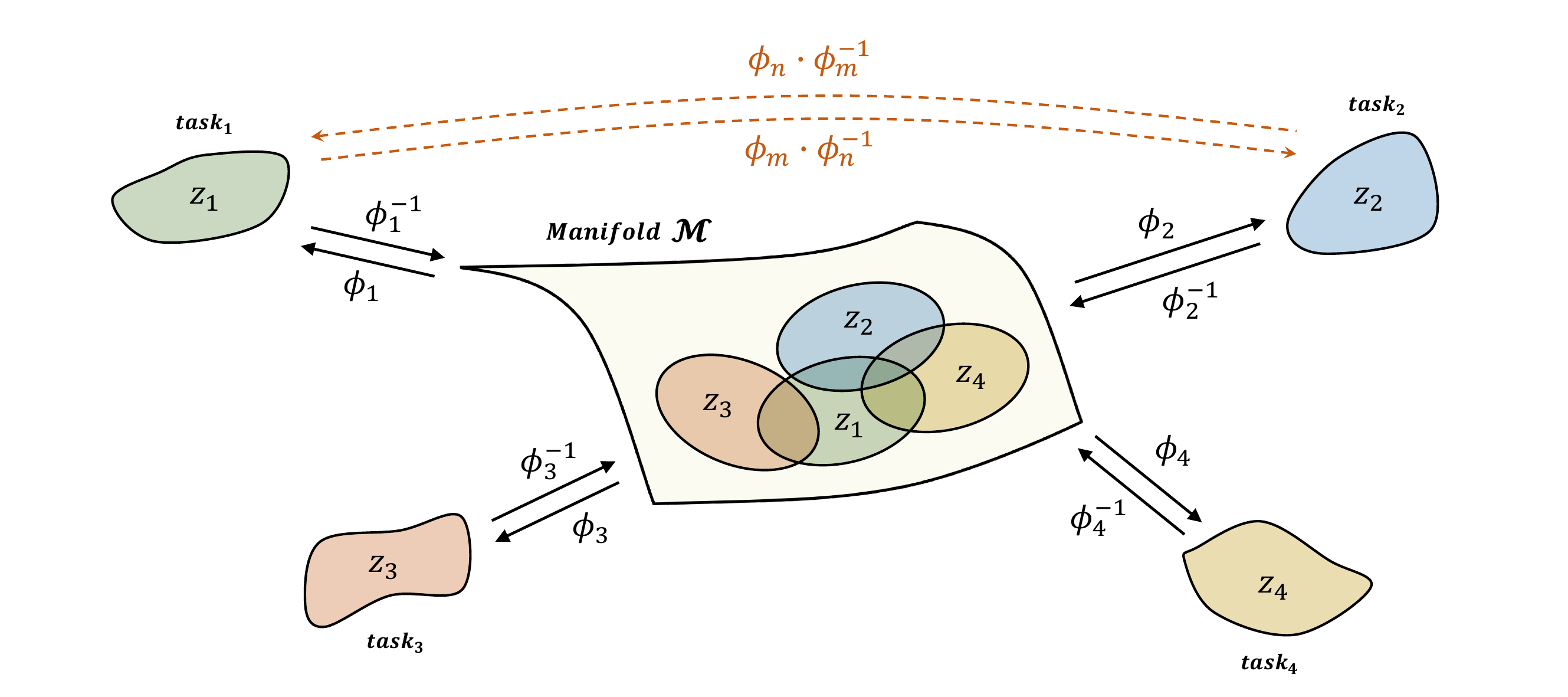}
\end{center}
\caption{Four different coordinate frames are demonstrated in the figure, with coordinate transformation maps to each pair of tasks. One can interpret each coordinate frame as task-specific coordinates and map them with transformation models. An arbitrary point in the overlapping region of the manifold can be transformed from one task coordinate to another by combining mapping functions $\phi$. Moreover, by introducing perturbation points, as demonstrated in the figure, one can define the distance between points to match the geometrical shape in the overlapping region.}
\label{fig:fig1}
\end{figure}

The fundamental concept remains unchanged. Since most molecular properties can be effectively computed from a common representation called SMILES \cite{weininger1988smiles}, it is natural to assume that there exists a common manifold for any tasks in molecular property prediction. Since this manifold is curved, imposing constraints to match the shapes of geometries of tasks requires a mapping from task coordinates to their corresponding locally flat frames. With multiple source tasks now present, it is mandatory to find mapping functions over task spaces for each one, as shown in Figure~\ref{fig:fig1}. This amplifies the leveraging effect of GATE, as mutual information now flows not only from one source task but also from multiple other sources.

We established an experimental setup based on the extended GATE algorithm with multiple molecular property prediction regression tasks from a number of different sources. We have shown that the extended GATE outperforms conventional multi-task learning schemes in terms of performance. Additionally, we conducted ablation tests to demonstrate that our algorithm is robust and reliable, even in extrapolation scenarios.

Our main contribution of the article is as follows.
\begin{itemize}
    \item We extend the GATE to encode multiple source tasks setup.
    \item Extension to multiple tasks provide a positive leveraging effect.
    \item Proposed model outperforms conventional method in multi-task molecular property setup.
\end{itemize}

\section{Multi-Task Extension of Geometrically Aligned Transfer Encoder}
Since the latent vector is believed to capture the essence of information for a given task, it is crucial to understand the geometrical characteristics of the latent spaces where the latent vector resides. If two different tasks share common factors in their property inference processes, then one may assume that the geometrical shapes of their latent spaces should be similar. Therefore, if one can align the geometrical shapes of tasks, mutual information will flow through mapping functions, thereby supporting the performance of the target task.

Here, we utilize the GATE algorithm and aim to extend its architecture to accommodate multiple source tasks.\footnote{For basic assumptions and detailed explanation of GATE, refer to \cite{ko2023geometrically}.}

\begin{figure}[t!]
\begin{center}
\includegraphics[width=0.92\linewidth]{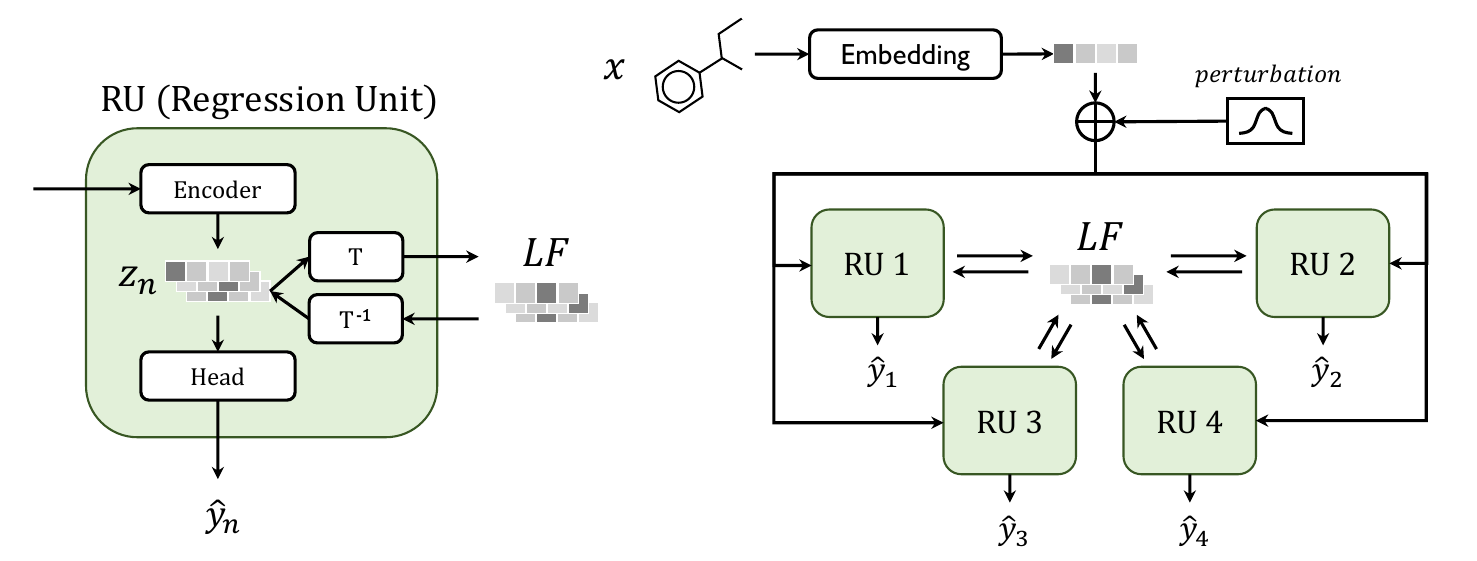}
\end{center}
\caption{Schematic diagram for the Extended GATE algorithm. The algorithm consists of a number of Regression Units. Each Regression Unit corresponds to an individual task. The universal manifold covers the entire coordinate space of RU's, and the transformation model maps a vector from each RU to a locally flat frame on the universal manifold. One can take the reverse path from the manifold to reconstruct the original vector. Furthermore, one can also transfer a vector to another RU coordinate by utilizing a different task's inverse transformation module.}
\label{fig:fig2}
\end{figure}

In Figure~\ref{fig:fig2} we first take an input SMILES and embed it into the corresponding vector. After embedding, latent space is formulated by encoders, which consist of DMPNN\cite{dmpnn} and MLP layers. The latent vector is fed into task-corresponding heads for inference properties. Here we utilize MSE for basic regression loss in the training scheme as follows:
\begin{gather}
    l_{\mathrm{reg}} = \frac{1}{N}\sum_{i}^{N}(y_i - \hat{y}_i)^2
\end{gather}
Where $N$, $y_i$, and $\hat{y}_i$ represent the number of data points, target, and predicted value, respectively. The difference now is that there exist multiple tasks, hence, there are also multiple instances of the regression loss.

To align the geometrical shapes of tasks, it is necessary to establish a mapping relation between the latent space and the locally flat frame of the universal manifold. The coordinate mapping can be induced by a Jacobian at an arbitrary point: %\footnote{Explicit derivation is in the appendix.}
\begin{gather}
    z'^i \equiv \sum_j \frac{\partial z'^i}{\partial z^j} z^j
\end{gather}
The model should always be able to differentiate in order to learn via gradient descent scheme. Hence, we design mapping function with autoencoder model. The encoder indicates mapping from latent space to universal manifold and decoder indicates mapping other way around.
\begin{gather} \label{transform1}
    z'_\alpha = \mathrm{Transfer}_{\alpha\rightarrow LF}(z_\alpha) \qquad    \hat{z}_\alpha = \mathrm{Transfer}^{-1}_{\phantom{-1}LF \rightarrow \alpha}(z'_\alpha)\\
    \label{transform2} z'_t = \mathrm{Transfer}_{t\rightarrow LF}(z_t) \qquad    \hat{z}_t = \mathrm{Transfer}^{-1}_{\phantom{-1}LF \rightarrow t}(z'_t)
\end{gather}
Where $t$ and $\alpha$ indicate target task and source number of task respectively. If there are $k$ numbers of source tasks, Greek alphabet runs from $1 \sim k$ and numbers indicate source task number. For instance, $\mathrm{Transfer}_{t}(z_t)$ means transformation from target latent to universal manifold and $\mathrm{Transfer}_{5}(z_5)$ means transformation from source task number $5$ to universal manifold.
We indeed utilize MSE loss for the autoencoder which consists of transfer and its inverse modules.
\begin{equation}
    l_{\mathrm{auto}} = \sum_\alpha \mathrm{MSE}(z_\alpha, \hat{z}_\alpha)
\end{equation}
Now everything is set to match geometrical shapes of latent spaces. Since encoder maps latent vector on latent space to locally flat frame on universal manifold, it is straight forward to impose a constraint that matches latent vector from target task and source task.

To define the consistency loss, we should recall the definition of the transformation model from the equations mentioned in \ref{transform1} and \ref{transform2}. As depicted in the equations, $\mathrm{Model}{0 \rightarrow LF}$ and $\mathrm{Model}{\alpha \rightarrow LF}$ indicate a model from the target to the locally flat (LF) frame and from the source to the LF frame, respectively. Here, we can impose a series of constraints to align the geometrical shapes from the source and target. One of these constraints requires that the latent vectors from the source and target should have the same value on the universal manifold. This constraint is referred to as the consistency loss.
\begin{equation}
    l_{cons} = \sum_\alpha \mathrm{MSE}(z'_\alpha, z'_t)
\end{equation}
This loss equalizes target latent and source latent vector in a locally flat frame on universal manifold. The latent spaces also aligned by latent vectors. Furthermore, one can induce another form of constraint to maximize the alignment of latent spaces.
\begin{gather}
    z'_\alpha = \mathrm{Transfer}_{\alpha\rightarrow LF}(z_\alpha) \qquad    \hat{z}_{\alpha \rightarrow t} = \mathrm{Transfer}^{-1}_{\phantom{-1}LF \rightarrow t}(z'_\alpha)
\end{gather}
The equation above illustrates the transformation of a latent vector from the source task to the target task. If the universal manifold is well-defined and both latent spaces from the source and target tasks are aligned properly, then a latent vector transformed from the source to the target task and a latent vector from the target task induced by the same SMILES input should always be the same. Hence, it is straightforward to imagine the specific form of the constraint which is written as follows.
\begin{equation}
    l_{map} = \sum_\alpha \mathrm{MSE}(y_t, \hat{y}_{\alpha \rightarrow t})
\end{equation}
Here, $y_t$ represents the label for the target predicted value, and $\hat{y}{s \rightarrow t}$ indicates the predicted value from $\hat{z}{\alpha \rightarrow t}$. The above loss ensures mutual information flow by aligning locally flat coordinates on the given latent vectors.

However, unfortunately, these constraints are insufficient to globally align latent spaces, as none of the introduced loss functions have locally bounded properties. Yet, it is necessary to impose another constraint that is not restricted to local properties.

In Riemannian geometry, it is common to attack geometric equations to find specific form of a metric of the given space. If one can find the explicit form of a metric, then the curvature of a given space can be identified, which can be utilized to understand the global characteristics of the space. Or, in other way around, if one has distance among points on a manifold, it is possible to find a metric from distance equation.
\begin{equation} \label{dist_curv}
    S^2 = \int_l \sum_\mu \sum_\nu g_{\mu\nu}dx^\mu dx^\nu
\end{equation}
However, in general, finding the analytic form of the metric is extremely complicated (or impossible). Therefore, we propose an idea to bypass this issue by utilizing the general mathematical characteristic of Riemannian geometry. In a curved space, distances between points are not intuitive and simple to compute. The metric is necessary to find finite distances. However, there is a wonderful invariance known as diffeomorphism in Riemannian manifolds. This invariance guarantees the freedom to fix coordinates by transformations induced by the Jacobian of a vector. And it is well-known that a locally flat frame is always possible to find around a given vector on a manifold. The locally flat frame, by its nature, is flat around the infinitesimal boundary of a vector. Therefore, the distance equation can now be reduced to a simpler form in local boundaries.
\begin{equation}
    S^2 = \int_l \sum_\mu \sum_\nu g_{\mu\nu}dx^\mu dx^\nu = \int_l \sum_\mu \sum_\nu \eta_{\mu\nu}dx^\mu dx^\nu = \int_a^b dx^2
\end{equation}
Here, $a$ indicates a given latent vector and $b$ is a perturbation around vector $a$. If this perturbation is infinitesimal, the distance between vector and its perturbation can be simplified as follows.
\begin{equation}
    S  =  |b - a|
\end{equation}

Now, for a given SMILES input and its infinitesimal perturbations, the latent vectors from the source and target tasks can be transformed into a vector on a universal manifold where the locally flat frame resides. One can compute distances between the latent vector and its perturbations from each task and require them to be the same. By doing so, the locally flat latent spaces will align together on a universal manifold and cover the overlapping region smoothly. Then, the mutual information can naturally be transferred from one to another, and the extrapolation performance of the model will be boosted by source data. In an abstract form, the distance loss can be expressed as follows.
\begin{equation}
    l_{dis} = \frac{1}{M}\sum_\alpha C_\alpha\sum_{i}^M \mathrm{MSE}(s^{i}_\alpha, s^{i}_t)
\end{equation}
Where $M$ is the number of perturbations, $C_\alpha$ is the given distance ratio for source to target, and $s^{i}_{s}$ is the displacement between pivot data points and their perturbations.
\begin{gather}
    s^{i}_\alpha \equiv |(z'_\alpha) - (z'^i_\alpha)| \qquad s^{i}_t \equiv |(z'_t) - (z'^i_t)| \\
    z'^i_\alpha = \mathrm{Transfer}_{\alpha\rightarrow LF}(\mathrm{Encoder}_\alpha(x^i))\\
    z'^i_t = \mathrm{Transfer}_{t\rightarrow LF}(\mathrm{Encoder}_t(x^i))   
\end{gather}
Here $x^i$ denotes $i$th perturbation of embedded input $x$, and $\mathrm{Encoder}_\alpha$ and $\mathrm{Encoder}_t$ are encoder parts of source and target model respectively. Finally, by gathering all losses with individual hyperparameters, we define the complete form of the loss function used in the extended GATE algorithm.
\begin{equation}
    l_{tot} = l_{reg} + \alpha l_{auto} + \beta l_{cons} + \gamma l_{map} + \delta l_{dis}
\end{equation}
Hyperparameters play a crucial role in weighted summation parameters, and by tuning them sophisticatedly, the model's performance will reach its peak. In most cases, many hyperparameters are sufficient to be set to a trivial number like $1$, but for parameters $\gamma$, $\delta$, and $C_\alpha$, it is worthwhile to tune them for optimal model performance. However, finding the right combinations of parameters can be challenging due to the immense search space. In such cases, we can rely on scientific knowledge to guide us in tuning them. %\footnote{The specific combination of parameters will be depicted in experiments section}

%%%%%%%%%%%%%%%%%%%%%%%%%%%%%%%%%%%%%%%%%%%%%%%%%%%%%%%%%%%%%%%%%%%%%%%%%%%%
%%%%%%%%%%%%%%%%%%%%%%%%%%%%%%%%%%%%%%%%%%%%%%%%%%%%%%%%%%%%%%%%%%%%%%%%%%%%
%%%%%%%%%%%%%%%%%%%%%%%%%%%%%%%%%%%%%%%%%%%%%%%%%%%%%%%%%%%%%%%%%%%%%%%%%%%%

\section{Experiments}
\subsection{Experimental Setup}

  A total of 10 datasets curated from five different sources named PubChem\cite{10.1093/nar/gkac956}, Ochem\cite{sushko2011online}, CCDDS, Yaws Handbook, and Jean-Claude Bradley were used for these experiments. We prepared the training and test sets by splitting each dataset according to the scaffold of the molecular structure\cite{bemis1996properties}. A single NVIDIA A40 was used for every experiment, and four-fold cross-validation setting with uniform sampling and a separated test set was used for the default setup. For all experiments, we consistently used the same architecture for encoders and heads.
  
  % Detailed experimental setup and information about datasets can be found in the appedix.

  % Every experiment is tested in a four-fold cross-validation setting with uniform sampling for accurate evaluation, and a single NVIDIA A40 is used for the experiments. For the evaluation, we compare the performance of GATE against that of single task learning (STL), MTL, KD, global structure preserving loss based KD (GSP-KD)~\cite{joshi2022representation}, and transfer learning (retrain all or head network only). We used the same architecture for encoders and heads in both the baselines and our model for all experiments. More detailed experimental setups can be found in Appendix. We performed experiments with a total of 23 target and source task pairs, respectively.

\subsection{Effect of multitask extension from two-task GATE to three-task GATE}

\begin{figure}[t!]
\begin{center}
\includegraphics[width=1\linewidth]{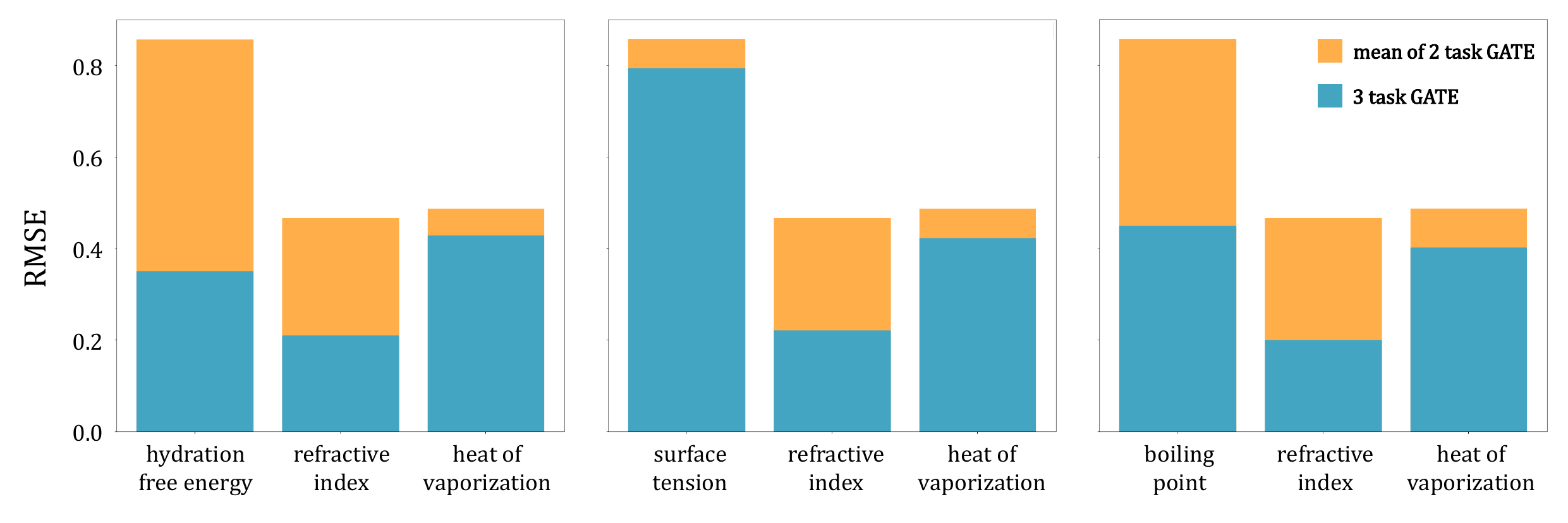}
\end{center}
\caption{Regression performance of three-task GATE and two-task GATE in root mean square error (RMSE). For evaluating regression performance of two-task GATE, all three possible pairs of three tasks were trained separately and averaged.}
\label{fig:fig3}
\end{figure}

  We first compared the regression performance of three-task GATE and two-task GATE to assess the impact of multitask extension. In each experiment, we used refractive index and heat of vaporization as pivot tasks and selected an additional task to constitute three tasks. Overall three set of experiments were performed using hydration free energy, surface tension or boiling point as an additional tasks respectively. To assess the regression performance of the two-task GATE, we separately trained and averaged all three possible combinations of the three tasks.
  
  As depicted in Figure~\ref{fig:fig3}, the results demonstrate a clear synergy effect among the three tasks. Across all three experiment sets, there is a consistent reduction in the root mean square error (RMSE) of the three-task GATE compared to the two-task GATE, even when different additional tasks are included in the sets. This result indicates that the prediction performance of molecular properties can be enhanced by incorporating suitable auxiliary tasks, and this synergy effect can be achieved through the proposed multitask extension of the GATE.

\subsection{Regression performance of many-task GATE}

\begin{table}[!ht]
\caption{Regression performance of 10-task GATE, MTL, and STL in Pearson correlation.}
\label{corr}
\begin{center}
    \begin{tabular}{|c|c|c|c|}
    \hline
        \textbf{Tasks} & \textbf{GATE} & \textbf{MTL} & \textbf{STL} \\ \hline \hline
        Parachor & 0.9303 & \textbf{0.9359} & 0.9287  \\ \hline
        Surface Tension & \textbf{0.8256} & 0.8195 & 0.7171  \\ \hline
        Dielectric Constant & 0.9067 & 0.9099 & \textbf{0.9216}  \\ \hline
        Hydration Free Energy & \textbf{0.9446} & 0.9409 & 0.9414  \\ \hline
        Viscosity & \textbf{0.9272} & 0.8952 & 0.8967  \\ \hline
        Boiling Point & 0.8930 & \textbf{0.9076} & 0.8847  \\ \hline
        Refractive Index & \textbf{0.9795} & 0.9781 & 0.9761  \\ \hline
        Density & \textbf{0.8518} & 0.8512 & 0.8237  \\ \hline
        Melting Point & 0.8715 & 0.8714 & \textbf{0.8901}  \\ \hline
        Heat of Vaporization & \textbf{0.9045} & 0.9018 & 0.8618  \\ \hline \hline
        No. 1st & \textbf{6} & 2 & 2  \\ \hline
    \end{tabular}
\end{center}
\end{table}

  To assess the effectiveness of GATE for multitask learning, we also compared the regression performance of the many-task GATE with that of classical multitask learning (MTL) techniques and single task learning (STL). As shown in Table~\ref{corr}, Pearson correlation of GATE outperforms MTL and STL for 6 out of 10 tasks, whereas MTL and STL perform best for only 2 tasks each.

  The advantage of the GATE for many-task setup is even more clearly shown in Table~\ref{percent}. Table~\ref{percent} shows percentage of improvement on regression performance of GATE and MTL compared to STL. As shown in the table, in many cases, multi-task setup enhances regression performance, but in some cases, it can actually reduce regression performance. This decline in performance can be attributed to negative transfer of undesired interfering information among the tasks. As evident from the table, GATE shows reduction of performance in only two tasks, while classical MTL exhibits performance decrease in four tasks out of ten tasks. This difference extends to more than 3.5 percent for viscosity, indicating that GATE significantly improves regression performance by a large margin, while MTL harms the regression. 

\begin{table}[!ht]
\caption{Relative improvement of the regression performance of 10-task GATE and MTL over STL in percent.}
\label{percent}
\begin{center}
    \begin{tabular}{|c|c|c|c|}
    \hline
        \textbf{Tasks} & \textbf{GATE} & \textbf{MTL} \\ \hline \hline
        Parachor & 0.18 & 0.78  \\ \hline
        Surface Tension & 15.13 & 14.28  \\ \hline
        Dielectric Constant & \textbf{\textcolor{red}{-1.61}} & \textbf{\textcolor{red}{-1.26}}  \\ \hline
        Hydration Free Energy & 0.33 & \textbf{\textcolor{red}{-0.06}}  \\ \hline
        Viscosity & \textbf{3.40} & \textbf{\textcolor{red}{-0.16}}  \\ \hline
        Boiling Point & 0.94 & 2.59  \\ \hline
        Refractive Index & 0.35 & 0.21  \\ \hline
        Density & 3.40 & 3.33  \\ \hline
        Melting Point & \textbf{\textcolor{red}{-2.09}} & \textbf{\textcolor{red}{-2.09}}  \\ \hline
        Heat of Vaporization & 4.96 & 4.64  \\ \hline
    \end{tabular}
\end{center}
\end{table}
  
  The result is well aligned with the experiments on stability of the latent spaces introduced in the original GATE paper\cite{ko2023geometrically}, which showed that the latent space of GATE exhibits relatively stable characteristics compared to that of MTL. Because the GATE is more resilient to interfering information, it exhibits more robust regression performance in a multi-task setup involving numerous tasks, where there is complex information exchange among the tasks.  

%%%%%%%%%%%%%%%%%%%%%%%%%%%%%%%%%%%%%%%%%%%%%%%%%%%%%%%%%%%%%%%%%%%%%%%%%%%%
%%%%%%%%%%%%%%%%%%%%%%%%%%%%%%%%%%%%%%%%%%%%%%%%%%%%%%%%%%%%%%%%%%%%%%%%%%%%
%%%%%%%%%%%%%%%%%%%%%%%%%%%%%%%%%%%%%%%%%%%%%%%%%%%%%%%%%%%%%%%%%%%%%%%%%%%%

\section{Discussion}
The original GATE algorithm interprets the latent space as a curved space and utilizes the mathematical concept of differential geometry, particularly Riemannian manifolds. Since the mathematical concept of GATE is not restricted to the two-task case, it is straightforward to generalize the algorithm to cover multiple source tasks without loss of generality. In this work, we designed the mathematical notion of the extended GATE with newly introduced hyperparameters and extended losses, and we have demonstrated the superior performance of the model using numerous open database datasets.

While our model often outperforms conventional setups, there are several areas for improvement. First, the model's computational complexity grows significantly with the number of source tasks. Since the distance and mapping losses must be computed for every pair of source and target tasks, the complexity is on the order of $\mathcal{O}(N^2)$. Therefore, compactifying the model architecture is one research direction to explore.

Second, the distance loss can potentially be omitted if one can directly calculate the curvature of the space by finding the analytic form of the metric tensor. While this is normally impossible, by utilizing the notion of operator learning, it can be achieved. After specifying the form of the metric tensor, one can pre-calculate the Ricci scalar of the space in advance. By matching the Ricci scalar from source and target spaces, the distance loss can be omitted and replaced. This idea can encode geometric information not restricted to local geometry but global, potentially improving GATE's performance and robustness even further.

%%%%%%%%%%%%%%%%%%%%%%%%%%%%%%%%%%%%%%%%%%%%%%%%%%%%%%%%%%%%%%%%%%%%%%%%%%%
%%%%%%%%%%%%%%%%%%%%%%%%%%%%%%%%%%%%%%%%%%%%%%%%%%%%%%%%%%%%%%%%%%%%%%%%%%%
% \begin{ack}
% Sung Moon Ko wants to thank to Yoonji Suh for helpful discussions for this work.
% \end{ack}

%%%%%%%%%%%%%%%%%%%%%%%%%%%%%%%%%%%%%%%%%%%%%%%%%%%%%%%%%%%%%%%%%%%%%%%%%%%
%%%%%%%%%%%%%%%%%%%%%%%%%%%%%%%%%%%%%%%%%%%%%%%%%%%%%%%%%%%%%%%%%%%%%%%%%%%

\bibliographystyle{unsrt}
\bibliography{arxiv_GATE}

\end{document}